\documentclass{article}

\pdfoutput=1

\usepackage{microtype}
\usepackage{graphicx}
\usepackage{subfigure}
\usepackage{booktabs} 
\usepackage{eufrak}
\usepackage{amsmath}

\usepackage{hyperref}

\usepackage[accepted]{icml2019_phys4dl}

\icmltitlerunning{Finite size corrections for NN Gaussian processes}

\newcommand{\measuredslope}{-1.07}

\begin{document}

\twocolumn[
\icmltitle{Finite size corrections for neural network Gaussian processes}




\begin{icmlauthorlist}
\icmlauthor{Joseph M.~Antognini}{whisper}
\end{icmlauthorlist}

\icmlaffiliation{whisper}{Whisper AI, San Francisco, California}

\icmlcorrespondingauthor{Joseph Antognini}{joe.antognini@gmail.com}

\icmlkeywords{Gaussian process, renormalization group}

\vskip 0.3in
]



\printAffiliationsAndNotice{}  

\begin{abstract}

    There has been a recent surge of interest in modeling neural networks (NNs) as Gaussian
    processes.  In the limit of a NN of infinite width the NN becomes equivalent to a Gaussian
    process.  Here we demonstrate that for an ensemble of large, finite, fully connected networks
    with a single hidden layer the distribution of outputs at initialization is well described by a
    Gaussian perturbed by the fourth Hermite polynomial for weights drawn from a symmetric
    distribution.  We show that the scale of the perturbation is inversely proportional to the
    number of units in the NN and that higher order terms decay more rapidly, thereby recovering the
    Edgeworth expansion.  We conclude by observing that understanding how this perturbation changes
    under training would reveal the regimes in which the Gaussian process framework is valid to
    model NN behavior.

\end{abstract}

\section{Introduction}
\label{sec:introduction}

Today it is well known that there is a deep connection between modern, highly overparameterized
neural networks (NNs) and Gaussian processes.  A foundational result in the field by \citet{neal96}
and \citet{williams97} demonstrated that a randomly initialized NN with a single hidden layer of
infinite width is identical to a Gaussian process so long as the weights of the NN are drawn from a
distribution with finite variance.  Although the covariance between different hidden units of the NN
is zero, these works showed that the covariance between a single hidden unit with different inputs
is non-zero, thereby making learning possible.

Although \citet{neal96} and \citet{williams97} only studied the case of NNs with a single hidden
layer, several recent works have extended this insight to show that NNs with multiple, possibly
convolutional, layers of infinite width are also Gaussian processes \citep{lee2017deep,
matthews2018gaussian, novak2019bayesian, garriga2018deep}.  And while these works only studied NNs
at initialization, \citet{jacot2018neural} showed that gradient descent on a NN corresponds to
applying a tangent kernel to an equivalent Gaussian process, and \citet{arora2019exact} has recently
weakened the conditions on this proof.  \citet{lee2019wide} empirically showed that application of
the tangent kernel closely match the predicted dynamics of a linearized deep NN.

Although NNs have grown dramatically in size, they have remained frustratingly finite.  In practice,
most practitioners tend to choose widths in the range 128--1024.  Especially as deep learning models
have moved onto mobile devices there has been substantial effort into compressing NNs so that they
can perform inference quickly and efficiently on low power devices
\citep[e.g.,][]{hinton2015distilling, han2015learning, howard2017mobilenets, sandler2018mobilenetv2,
howard2019searching, tan2019efficientnet}.

In this paper we attempt to bridge the divide between the infinite NNs of theory and the merely
large NNs of practice.  We show that the Edgeworth expansion is a useful tool to describe the
distribution of NN outputs for large NN ensembles.  In particular, the distribution of outputs for
an ensemble of large, finite NNs is not Gaussian, but is instead a Gaussian perturbed by the fourth
Hermite polynomial, and the magnitude of this perturbation is inversely proportional to the number
of hidden units.  The author believes that understanding the stability of this perturbation under
training will be a useful method to assess the relevance of the NN Gaussian process framework to
study the long-term dynamics of NN training.

\section{Derivation of the Gaussian process limit using the renormalization group}
\label{sec:derivation}

Let us consider a NN with a single hidden layer consisting of $N$ units.  For simplicity of notation
we shall restrict ourselves to the case where both the input, $x$, and the output, $y$, are
one-dimensional.  The more general case of a multi-dimensional input does not change the derivation,
and as \citet{neal96} notes, when the output is multi-dimensional the different output dimensions
are independent.  We write the input-to-hidden weights as $u_i$, and the hidden-to-output weights as
$v_i$.  We will omit bias terms since biases are commonly initialized with zeros.  Their inclusion
clutters the math but does not change the results.  The hidden activations are given by
\begin{equation}
    h_i = f(u_i x),
\end{equation}
where $f$ is the activation function.  The output of the NN is given by
\begin{equation}
    \label{eq:output}
    y = \sum_{i = 1}^N v_i h_i.
\end{equation}
Let us suppose that the weights are initialized by an i.i.d.~sample from a probability distribution
(not necessarily the same for different layers).  At initialization the parameters $\textbf{u}$,
$\textbf{v}$, and $\textbf{b}$ are generally sampled independently from a set of probability
distributions with finite variance.\footnote{An important exception is orthogonal initialization
\citep{saxe2013exact} which places an orthogonality constraint on the weight matrices thereby
causing the individual parameters to lose their independence.} Most practitioners today use Glorot
uniform initialization \cite{glorot2010understanding} and this is the default in popular frameworks
like Tensorflow \citep{abadi2016tensorflow} and PyTorch \citep{paszke2017automatic}.

As \citet{neal96} observes, because the $u_i$ are all independent of one another, the covariance
between any two $h_i$ must be zero for fixed $x$.\footnote{And, because the $v_i$ are all
independent of one another, the covariance between two different output dimensions must be zero for
fixed $x$ as well (thereby justifying our study of the case of a single output dimension).  Since
this is a simply a statement about expectations and does not require taking the limit $N \to \infty$
it holds for NNs of both infinite and finite width.  The $N \to \infty$ limit was only required by
\citet{neal96} to guarantee a Gaussian distribution; it does not change the covariance, assuming
that the weights are drawn from a distribution that scales inversely with $N$ (which is required to
keep the output finite in the limit of infinite $N$).}  However, the covariance of a fixed hidden
unit for two \emph{different} inputs is, in general, non-zero.  This covariance can be written as a
kernel, $C(x, x^{\prime})$, which in the Gaussian process limit, expresses the entire state of the
Gaussian process.  Note however that the calculation of the covariance is independent of the number
of hidden units.  The requirement that the number of units in the hidden layer go to infinity is
necessary for the distribution of outputs to be Gaussian.  If the number of hidden units is finite,
the covariance remains the same, but the output distribution may be different.

To determine the effect of a large, but finite, number of units in the hidden layer we use the
renormalization group to recover the Edgeworth expansion.  There are a number of works which derive
the central limit theorem and Edgeworth expansion using the renormalization group
\citep[e.g.,][]{anshelevich1999linearization, jona2001renormalization, calvo2010generalized}, and
here we follow an approach in \citet[pp.~291--3]{sethna2006statistical}.  The renormalization group
analysis consists of four steps:

1. Coarse-grain.

2. Renormalize.

3. Find the fixed point of the renormalization transformation.

4. Linearize about this fixed point.

\subsection{Coarse-graining}

To coarse-grain we observe that the output of the NN is given by the sum over all hidden units in
Eq.~\ref{eq:output}.  For a fixed input, the value of each activation may be considered to be a
sample from some probability distribution $p_h(h; x)$.  Here we write $h$ as an argument of the
probability distribution function and treat the input $x$ as a parameter since it affects the shape
of the distribution but is not a random variate itself.  Similarly, the value of $v_i$ is given by a
sample from $p_v(v; x)$.  Since the samples $u_i$ are independent, the $h_i$ are independent as well,
and the output is given by the sum of the products of samples from the two probability distributions
$p_h(h; x)$ and $p_v(v; x)$.  This product of $h_i$ and $v_i$ can be considered to be a sample from a
third probability distribution $p_{hv}(hv; x)$.

We now replace each pair of hidden units with a single hidden unit:
\begin{equation}
    (hv)^{\prime}_i = (hv)_{2i} + (hv)_{2i + 1}.
\end{equation}
The probability distribution of the transformed $(hv)^{\prime}$ is simply the convolution of the
probability distribution $p_{hv}(hv)$ with itself.  It is easier to represent this transformation in
Fourier space because the convolution becomes multiplication:
\begin{equation}
    \widetilde{p}^{\prime}_{hv}(k; x) = \widetilde{p}_{hv}(k; x)^2,
\end{equation}
where we use a tilde to represent the Fourier transformed distribution, and $k$ to represent the
frequency domain of the product $hv$.

\subsection{Renormalization}

In the next step we renormalize the transformed probability distribution so that it becomes
self-similar to the original probability distribution.  Specifically, when we add two random
variates, the standard deviation of the result is larger by the factor $\sqrt{2}$.  We therefore
need to rescale the new probability distribution so that it has the same variance as the original.
(We are assuming that the probability distributions we are working with are centered so the mean
does not need to be transformed.)  The rescaled probability distribution is then
$\sqrt{2}p^{\prime}_{hv}(\sqrt{2} hv; x)$, where the prefactor is required to normalize the rescaled
probability distribution.  The final renormalization operator is therefore
\begin{eqnarray}
    \mathfrak{R}[p_{hv}(hv; x)] & \equiv & 2 p_{hv}(\sqrt{2} hv; x) * p_{hv}(\sqrt{2} hv; x) \\
    & = & \mathcal{F}^{-1}[\widetilde{p}_{hv}(k / \sqrt{2}; x)^2].
\end{eqnarray}

\subsection{The fixed point of the renormalization transformation}

At the fixed point of the renormalization transformation, successive applications of the
transformation do not change the distribution, so we have
\begin{equation}
    p^*_y(y; x) = \mathfrak{R}[p^*_y(y; x)],
\end{equation}
where we use an asterisk to represent the fixed point and we now identify the sum of the $h_i v_i$
from the repeated application of the transformation as the NN output $y$.  Taking the Fourier
transform we have
\begin{equation}
    \widetilde{p}^*_y(k; x) = \widetilde{p}^*_y(k / \sqrt{2}; x)^2.
\end{equation}
The solution to this equation is the Gaussian distribution,
\begin{equation}
    \widetilde{p}^*_y(k; x) = \mathcal{N}(k; 0, \sigma^2) \equiv 
    \frac{1}{\sqrt{2 \pi} \sigma} e^{-k^2 / (2 \sigma^2)},
\end{equation}
where $\sigma$ is the standard deviation of the original distribution and is a function of the input
$x$.  (Due to the renormalization transformation we constrain $\sigma$ to remain fixed for fixed
$x$.)  This is just a restatement of the fact that the Gaussian distribution is the stable
distribution for the family of distributions with finite variance \citep[\textsection
VIII.4]{feller66}.

\subsection{Linearization about the fixed point}

Let us now consider a probability distribution which is \emph{close} to a Gaussian distribution, but
is not exactly Gaussian.  We can represent this distribution as $p_y(y) = p^*_y(y) + \epsilon
\phi(y)$ for some small $\epsilon$ and arbitrary function, $\phi(y)$.  We can then linearize the
renormalization transformation by finding its eigenvalues and eigenfunctions.  These must satisfy
the relationship
\begin{equation}
    \mathfrak{R}[p^*_y(y; x) + \epsilon \phi(y; x)] \simeq p^*_y(y; x) +
    \sum_{n=0}^{\infty} \lambda_n \epsilon \phi_n(y; x),
\end{equation}
where we drop terms of order $\epsilon^2$ and higher.  Substituting the renormalization
transformation and taking the Fourier transform, we find
\begin{equation}
    \widetilde{\phi}_n(k; x) = \frac{1}{\lambda_n \sigma} \sqrt{\frac{2}{\pi}} e^{-k^2 / 4 \sigma}
    \widetilde{\phi}_n \left( \frac{k}{\sqrt{2}}; x \right).
\end{equation}
The set of eigenfunctions that satisfy this relationship in Fourier space is given by
\begin{equation}
    \widetilde{\phi}_n(k; x) = (ik)^n \mathcal{N}(k; 0, \sigma^2).
\end{equation}
Taking the inverse Fourier transform we find that the eigenfunctions are Hermite polynomials
multiplied by a Gaussian:
\begin{equation}
    \label{eq:eigenfuncitions}
    \phi_n(y) = H_n(y) \mathcal{N}(y; 0, \sigma^2),
\end{equation}
where $H_n(x) \equiv (x - D)^n \cdot 1$, with $D$ being the differential operator.  From these
eigenfunctions we find that the eigenvalues are
\begin{equation}
    \lambda_n = 2^{1 - n/2}.
\end{equation}
Note that the first two eigenvalues are relevant, and the third is marginal.  This is a consequence
of the fact that as the width of the NN tends to infinity, the resulting output distribution must
remained normalized and have a fixed mean and standard deviation.  The rest of the eigenvalues are
irrelevant, however, due to the fact that in the infinite width limit, the higher order moments must
tend to the values of a Gaussian.

Now, since we have $N$ units in the hidden layer of the NN, we will need to apply the
renormalization transformation $\log_2 N$ times.  The eigenvalues for the repeated transformation
will therefore be 
\begin{equation}
    \label{eq:repeated_eigenvalue}
    \lambda_n^{\log_2 N} = N^{1 - n/2}.
\end{equation}

We can now write out the probability distribution for the NN output as
\begin{eqnarray}
    p_y(y) & = &
      p^*_y(y) + c_3(x) \lambda_3^{\log_2 N} \phi_3(hv; x) + \nonumber \\
      & & c_4(x) \lambda_4^{\log_2 N} \phi_4(y; x) + \cdots \\
    & = & \mathcal{N}(y; 0, \sigma^2) \times (1 + c_3(x) N^{-1/2} H_3(y) + \nonumber \\
    & & c_4(x) N^{-1} H_4(y) + \cdots),
\end{eqnarray}
where we explicitly represent the constants $c_i$ as being functions of the input $x$ (because they
are determined by the moments of the original probability distribution $p_y(y; x)$), and expect them
to be of order unity in general.  This result is a recovery of the Edgeworth expansion, but in this
case it is the expansion of a stochastic process rather than a probability distribution because it
is parameterized by the input, $x$ \citep[e.g.,][]{juszkiewicz+95}.\footnote{In the infinite width
limit the joint output distribution of two or more different inputs is given by a multivariate
Gaussian distribution.  In the finite width case we expect the joint distribution of two or more
different inputs to be given by the multivariate Edgeworth expansion, although we do not show it
here.  See \citet{sellentin+17} for a derivation of the multivariate Edgeworth expansion.}  Given
the distribution $p_y(y; x)$ the constants $c_i$ can be determined exactly from a more rigorous
derivation \citep[e.g.,][]{hansen06edgeworth}.  The Edgeworth expansion is an asymptotic series, so
for fixed $n$ the series converges as $N \to \infty$, but the series itself diverges as $n \to
\infty$.  For large $N$, then, we may drop the $c_5$ term and higher.

Now, it is generally the case that the probability distributions used to initialize NNs are
symmetric and so have zero skew.  This symmetry forces the first irrelevant eigenfunction (given by
the third Hermite polynomial) to make no contribution so that $c_3 = 0$.  This implies that for the
probability distributions typically used to initialize NNs we have
\begin{multline}
    \label{eq:perturbation}
    p_y(y) \simeq \frac{1}{\sqrt{2 \pi} \sigma} e^{-y^2 / 2 \sigma^2} \times \\
    \left[1 -
    \frac{c_4(x)}{N} \left( 3 - 6 \left(\frac{y}{\sigma}\right)^2 + \left( \frac{y}{\sigma}
    \right)^4 \right) \right].
\end{multline}

\section{Experiments}

\subsection{The perturbation from the Gaussian}

To verify the correctness of Eq.~\ref{eq:perturbation} we generate an ensemble of $10^8$ randomly
initialized NNs.  Each NN has a single hidden layer with 128 units and a ReLU activation.  The NNs
are initialized using the default \texttt{layers} behavior in Tensorflow with the weights drawn from
the Glorot uniform distribution and the biases set to zero.  We then observe the distribution of
outputs for a fixed input of $x = 1$.  The resulting distribution should be close to, but not
exactly, Gaussian.  To measure the deviation from Gaussianity we calculate the empirical cumulative
distribution function (CDF) of the outputs and subtract the CDF of a Gaussian distribution with the
same variance as the outputs ($1/64$ for this NN).

\begin{figure}
    \includegraphics[width=8cm]{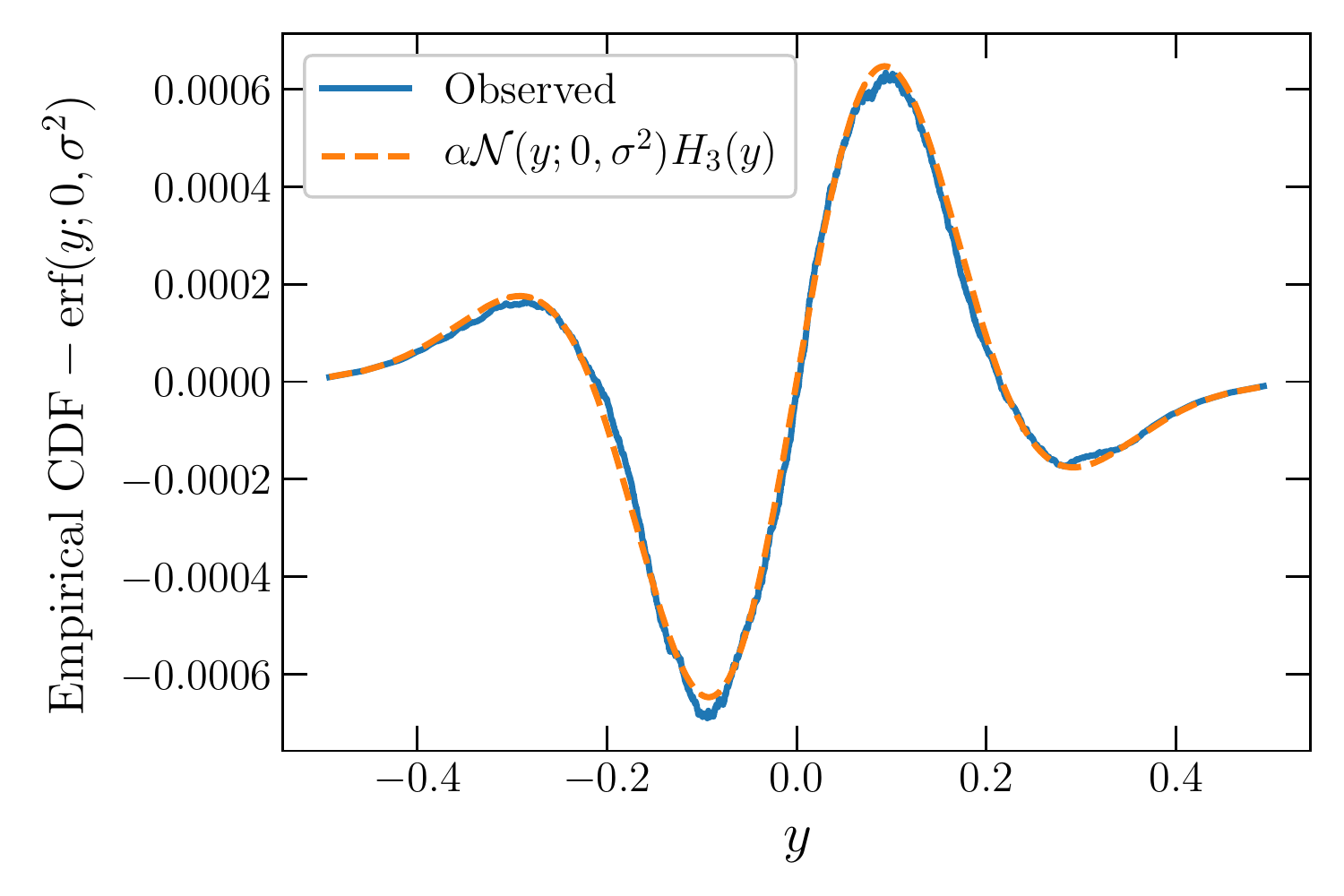}

    \caption{The difference between the empirical CDF of the output of an ensemble of randomly
    initialized NNs with the CDF of a Gaussian (solid blue line).  The predicted difference from
    Eq.~\ref{eq:cdf_diff} is shown in the dashed orange line.  The empirical CDF was calculated for
    $10^8$ NNs each with a single hidden layer consisting of 128 units for a fixed input of $x =
    1$.}

    \label{fig:cdf_diff}
\end{figure}

Calculating the CDF of $p_y(y)$ in Eq.~\ref{eq:perturbation} and subtracting off the CDF of a
normal distribution, we find that the difference is given by the third Hermite polynomial times a
Gaussian:
\begin{equation}
    \label{eq:cdf_diff}
    \int_{-\infty}^{y} p_y(y^{\prime}) - \mathcal{N}(y^{\prime}; 0, \sigma^2) \, dy^{\prime} = 
    \frac{c_4}{N} \mathcal{N}(y; 0, \sigma^2) H_3(y).
\end{equation}
We compare the empirical CDF with the predicted CDF in Fig.~\ref{fig:cdf_diff} and find excellent
agreement.  From this empirical CDF we measure $c_4 \approx 9.405$.

\subsection{The magnitude of the perturbation and $N$}

\begin{figure}
    \centering
    \includegraphics[width=8cm]{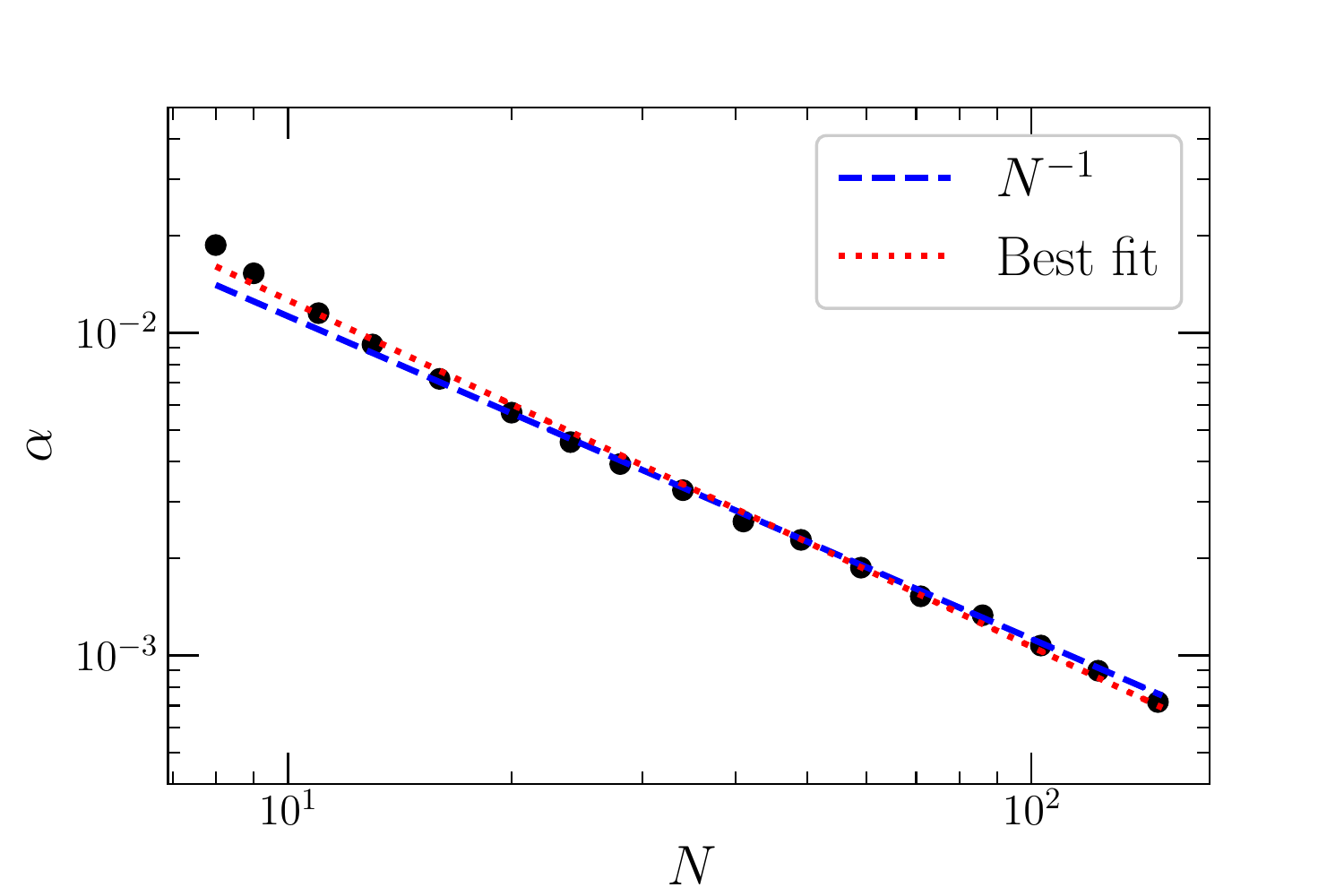}

    \caption{The scaling of the perturbation of the distribution from a Gaussian with the number of
    hidden units.  Each point is the best fit for $\alpha$ of the difference between the empirical
    CDF with the CDF of the third Hermite polynomial times a Gaussian (see Eq.~\ref{eq:scaling}).
    The predicted $N^{-1}$ scaling is shown with the dashed blue line and the best fit scaling
    $N^{\measuredslope}$ is shown with the dotted red line.}

    \label{fig:scaling}

\end{figure}

From Eq.~\ref{eq:perturbation} we expect the magnitude of the deviation from a Gaussian to scale
inversely with the number of hidden units, $N$.  To test this prediction we generate a set of
ensembles of randomly initialized NNs for a range of $N$.  We vary $N$ between 8 and 148 and for
each $N$ we use an ensemble of $10^7$ NNs.  As before, the weights are initialized from a Glorot
uniform distribution, the biases are set to zero, and a ReLU activation is used.  The input is fixed
to $x = 1$ and the distribution of output values is collected across the ensemble.  We then
calculate the difference between the empirical CDF and a Gaussian CDF (with the variance of the
Gaussian set to be the observed variance of the outputs).  This difference is expected to be the
third Hermite polynomial times a Gaussian times a scaling factor $\alpha$ and measure the best fit
for $\alpha$:
\begin{equation}
    \label{eq:scaling}
    \int_{-\infty}^{y} p_y(y^{\prime}) - \mathcal{N}(y^{\prime}; 0, \sigma^2) \, dy^{\prime} = 
    \alpha \mathcal{N}(y; 0, \sigma^2) H_3(y).
\end{equation}
The results for the measured values of $\alpha$ are shown in Fig.~\ref{fig:scaling}.  We find that
$\alpha$ scales approximately with the inverse of $N$ as expected, except for small values of $N$,
where the dependence is slightly steeper.   The best fit is $\alpha \propto N^{\measuredslope}$.

\section{Discussion}

\citet{jacot2018neural} has shown that if an infinitely wide NN is trained on a mean squared error
loss then the NN remains a Gaussian process throughout training.  However, NNs used in practice are
not infinitely wide and therefore are not Gaussian.  Nevertheless, this analysis shows that the
perturbation away from a Gaussian can be quantified and scales approximately inversely with $N$.  It
is therefore interesting to ask what happens to the perturbation when a NN is trained.  Does it
shrink over the course of training, stay the same magnitude, or increase?

In the first case we should expect that Gaussian processes will be a powerful framework to
understand NNs because even if a typical NN is not quite a Gaussian process at the beginning of
training, it will soon become one.  In the second case, Gaussian processes will still be a useful
framework, but with the understanding that the distributions are in fact perturbed Gaussian
distributions rather than exactly Gaussian.  If the final case holds, Gaussian processes will be a
useful framework to understand the early stages of training, but will become progressively worse.
Depending on how great the deviation from Gaussianity becomes, the Gaussian process framework may
fail to describe the training dynamics entirely at a certain point in training.  (If this occurs
then we would be able to identify the point at which this perturbation diverges as a phase
transition in the NN.)  Based on the empirical success of Gaussian processes at predicting the
trajectories of NNs of finite sizes in \citet{lee2019wide} it is the opinion of the author that the
output distribution is unlikely to stray too far from a Gaussian over the course of training.
However, rigorously proving this to be the case is a more complicated problem than analyzing the
perturbation from a Gaussian at initialization and is beyond the scope of this paper.

\section*{Acknowledgments}

The author is grateful to Michael Flynn, Kevin Quetano, Josh Schneier, and Richard Zhu for helpful
discussions and suggestions.

\bibliography{refs.bib}

\begin{thebibliography}{27}
\providecommand{\natexlab}[1]{#1}
\providecommand{\url}[1]{\texttt{#1}}
\expandafter\ifx\csname urlstyle\endcsname\relax
  \providecommand{\doi}[1]{doi: #1}\else
  \providecommand{\doi}{doi: \begingroup \urlstyle{rm}\Url}\fi

\bibitem[Abadi et~al.(2016)Abadi, Barham, Chen, Chen, Davis, Dean, Devin,
  Ghemawat, Irving, Isard, et~al.]{abadi2016tensorflow}
Abadi, M., Barham, P., Chen, J., Chen, Z., Davis, A., Dean, J., Devin, M.,
  Ghemawat, S., Irving, G., Isard, M., et~al.
\newblock Tensorflow: A system for large-scale machine learning.
\newblock In \emph{12th $\{$USENIX$\}$ Symposium on Operating Systems Design
  and Implementation ($\{$OSDI$\}$ 16)}, pp.\  265--283, 2016.

\bibitem[Anshelevich(1999)]{anshelevich1999linearization}
Anshelevich, M.
\newblock The linearization of the central limit operator in free probability
  theory.
\newblock \emph{Probability theory and related fields}, 115\penalty0
  (3):\penalty0 401--416, 1999.

\bibitem[Arora et~al.(2019)Arora, Du, Hu, Li, Salakhutdinov, and
  Ruosong]{arora2019exact}
Arora, S., Du, S.~S., Hu, W., Li, Z., Salakhutdinov, R., and Ruosong, W.
\newblock On exact computation with an infinitely wide neural net.
\newblock \emph{arXiv preprint arXiv:1904:11955}, 2019.

\bibitem[Calvo et~al.(2010)Calvo, Cuch{\'\i}, Esteve, and
  Falceto]{calvo2010generalized}
Calvo, I., Cuch{\'\i}, J.~C., Esteve, J.~G., and Falceto, F.
\newblock Generalized central limit theorem and renormalization group.
\newblock \emph{Journal of Statistical Physics}, 141\penalty0 (3):\penalty0
  409--421, 2010.

\bibitem[Feller(1966)]{feller66}
Feller, W.
\newblock \emph{An introduction to probability theory and its applications},
  volume~2.
\newblock John Wiley \& Sons, 1966.

\bibitem[Garriga-Alonso et~al.(2018)Garriga-Alonso, Aitchison, and
  Rasmussen]{garriga2018deep}
Garriga-Alonso, A., Aitchison, L., and Rasmussen, C.~E.
\newblock Deep convolutional networks as shallow gaussian processes.
\newblock \emph{arXiv preprint arXiv:1808.05587}, 2018.

\bibitem[Glorot \& Bengio(2010)Glorot and Bengio]{glorot2010understanding}
Glorot, X. and Bengio, Y.
\newblock Understanding the difficulty of training deep feedforward neural
  networks.
\newblock In \emph{Proceedings of the thirteenth international conference on
  artificial intelligence and statistics}, pp.\  249--256, 2010.

\bibitem[Han et~al.(2015)Han, Pool, Tran, and Dally]{han2015learning}
Han, S., Pool, J., Tran, J., and Dally, W.
\newblock Learning both weights and connections for efficient neural network.
\newblock In \emph{Advances in neural information processing systems}, pp.\
  1135--1143, 2015.

\bibitem[Hansen(2006)]{hansen06edgeworth}
Hansen, E.
\newblock Edgeworth expansions, 2006.
\newblock URL
  \url{http://web.math.ku.dk/~erhansen/bootstrap_05/doku/noter/Edgeworth_24_01.pdf}.

\bibitem[Hinton et~al.(2015)Hinton, Vinyals, and Dean]{hinton2015distilling}
Hinton, G., Vinyals, O., and Dean, J.
\newblock Distilling the knowledge in a neural network.
\newblock \emph{arXiv preprint arXiv:1503.02531}, 2015.

\bibitem[Howard et~al.(2019)Howard, Sandler, Chu, Chen, Chen, Tan, Wang, Zhu,
  Pang, Vasudevan, et~al.]{howard2019searching}
Howard, A., Sandler, M., Chu, G., Chen, L.-C., Chen, B., Tan, M., Wang, W.,
  Zhu, Y., Pang, R., Vasudevan, V., et~al.
\newblock Searching for mobilenetv3.
\newblock \emph{arXiv preprint arXiv:1905.02244}, 2019.

\bibitem[Howard et~al.(2017)Howard, Zhu, Chen, Kalenichenko, Wang, Weyand,
  Andreetto, and Adam]{howard2017mobilenets}
Howard, A.~G., Zhu, M., Chen, B., Kalenichenko, D., Wang, W., Weyand, T.,
  Andreetto, M., and Adam, H.
\newblock Mobilenets: Efficient convolutional neural networks for mobile vision
  applications.
\newblock \emph{arXiv preprint arXiv:1704.04861}, 2017.

\bibitem[Jacot et~al.(2018)Jacot, Gabriel, and Hongler]{jacot2018neural}
Jacot, A., Gabriel, F., and Hongler, C.
\newblock Neural tangent kernel: Convergence and generalization in neural
  networks.
\newblock In \emph{Advances in neural information processing systems}, pp.\
  8571--8580, 2018.

\bibitem[Jona-Lasinio(2001)]{jona2001renormalization}
Jona-Lasinio, G.
\newblock Renormalization group and probability theory.
\newblock \emph{Physics Reports}, 352\penalty0 (4-6):\penalty0 439--458, 2001.

\bibitem[{Juszkiewicz} et~al.(1995){Juszkiewicz}, {Weinberg}, {Amsterdamski},
  {Chodorowski}, and {Bouchet}]{juszkiewicz+95}
{Juszkiewicz}, R., {Weinberg}, D.~H., {Amsterdamski}, P., {Chodorowski}, M.,
  and {Bouchet}, F.
\newblock {Weakly nonlinear Gaussian fluctuations and the edgeworth expansion}.
\newblock \emph{The Astrophysical Journal}, 442:\penalty0 39--56, March 1995.
\newblock \doi{10.1086/175420}.

\bibitem[Lee et~al.(2017)Lee, Bahri, Novak, Schoenholz, Pennington, and
  Sohl-Dickstein]{lee2017deep}
Lee, J., Bahri, Y., Novak, R., Schoenholz, S.~S., Pennington, J., and
  Sohl-Dickstein, J.
\newblock Deep neural networks as gaussian processes.
\newblock \emph{arXiv preprint arXiv:1711.00165}, 2017.

\bibitem[Lee et~al.(2019)Lee, Xiao, Schoenholz, Bahri, Sohl-Dickstein, and
  Pennington]{lee2019wide}
Lee, J., Xiao, L., Schoenholz, S.~S., Bahri, Y., Sohl-Dickstein, J., and
  Pennington, J.
\newblock Wide neural networks of any depth evolve as linear models under
  gradient descent.
\newblock \emph{arXiv preprint arXiv:1902.06720}, 2019.

\bibitem[Matthews et~al.(2018)Matthews, Rowland, Hron, Turner, and
  Ghahramani]{matthews2018gaussian}
Matthews, A. G. d.~G., Rowland, M., Hron, J., Turner, R.~E., and Ghahramani, Z.
\newblock Gaussian process behaviour in wide deep neural networks.
\newblock \emph{arXiv preprint arXiv:1804.11271}, 2018.

\bibitem[Neal(1996)]{neal96}
Neal, R.~M.
\newblock Priors for infinite networks.
\newblock In \emph{Bayesian Learning for Neural Networks}, pp.\  29--53.
  Springer, 1996.

\bibitem[Novak et~al.(2019)Novak, Xiao, Bahri, Lee, Yang, Abolafia, Pennington,
  and Sohl-dickstein]{novak2019bayesian}
Novak, R., Xiao, L., Bahri, Y., Lee, J., Yang, G., Abolafia, D.~A., Pennington,
  J., and Sohl-dickstein, J.
\newblock Bayesian deep convolutional networks with many channels are gaussian
  processes.
\newblock In \emph{International Conference on Learning Representations}, 2019.
\newblock URL \url{https://openreview.net/forum?id=B1g30j0qF7}.

\bibitem[Paszke et~al.(2017)Paszke, Gross, Chintala, Chanan, Yang, DeVito, Lin,
  Desmaison, Antiga, and Lerer]{paszke2017automatic}
Paszke, A., Gross, S., Chintala, S., Chanan, G., Yang, E., DeVito, Z., Lin, Z.,
  Desmaison, A., Antiga, L., and Lerer, A.
\newblock Automatic differentiation in pytorch.
\newblock In \emph{NIPS 2017 Autodiff Workshop}, 2017.
\newblock URL \url{https://openreview.net/forum?id=BJJsrmfCZ}.

\bibitem[Sandler et~al.(2018)Sandler, Howard, Zhu, Zhmoginov, and
  Chen]{sandler2018mobilenetv2}
Sandler, M., Howard, A., Zhu, M., Zhmoginov, A., and Chen, L.-C.
\newblock Mobilenetv2: Inverted residuals and linear bottlenecks.
\newblock In \emph{Proceedings of the IEEE Conference on Computer Vision and
  Pattern Recognition}, pp.\  4510--4520, 2018.

\bibitem[Saxe et~al.(2013)Saxe, McClelland, and Ganguli]{saxe2013exact}
Saxe, A.~M., McClelland, J.~L., and Ganguli, S.
\newblock Exact solutions to the nonlinear dynamics of learning in deep linear
  neural networks.
\newblock \emph{arXiv preprint arXiv:1312.6120}, 2013.

\bibitem[{Sellentin} et~al.(2017){Sellentin}, {Jaffe}, and
  {Heavens}]{sellentin+17}
{Sellentin}, E., {Jaffe}, A.~H., and {Heavens}, A.~F.
\newblock {On the use of the Edgeworth expansion in cosmology I: how to foresee
  and evade its pitfalls}.
\newblock \emph{arXiv e-prints}, September 2017.

\bibitem[Sethna(2006)]{sethna2006statistical}
Sethna, J.
\newblock \emph{Statistical mechanics: entropy, order parameters, and
  complexity}, volume~14.
\newblock Oxford University Press, 2006.

\bibitem[Tan \& Le(2019)Tan and Le]{tan2019efficientnet}
Tan, M. and Le, Q.~V.
\newblock Efficientnet: Rethinking model scaling for convolutional neural
  networks.
\newblock \emph{arXiv preprint arXiv:1905.11946}, 2019.

\bibitem[Williams(1997)]{williams97}
Williams, C.~K.
\newblock Computing with infinite networks.
\newblock In \emph{Advances in neural information processing systems}, pp.\
  295--301, 1997.

\end{thebibliography}
\bibliographystyle{icml2019}

\end{document}